%% file: 0-main.tex
\documentclass{article}
\usepackage{caption} 
\usepackage{subcaption}

\usepackage{arxiv}
\usepackage[utf8]{inputenc} 
\usepackage[T1]{fontenc}    
\usepackage{hyperref}       
\usepackage{url}            
\usepackage{booktabs}       
\usepackage{amsfonts}       
\usepackage{amsmath}
\usepackage{nicefrac}       
\usepackage{blindtext}
\usepackage{titlesec}
\usepackage{microtype}      
\usepackage{fancyhdr}
\usepackage{lipsum}
\usepackage{graphicx}
\usepackage{enumitem}
\usepackage{soul}
\usepackage{listings}
\usepackage{multirow} 
\usepackage[table,xcdraw]{xcolor}
\usepackage{float}
\usepackage[none]{hyphenat}
\usepackage{fancyvrb,fvextra}
\usepackage{csquotes}
\usepackage{natbib}

\usepackage[
  separate-uncertainty = true,
  multi-part-units = repeat
]{siunitx}
\usepackage[normalem]{ulem}
\useunder{\uline}{\ul}{}

\graphicspath{ {./images/} }
\usepackage{authblk}
\author[1]{Pawitsapak Akarajaradwong}
\author[1]{Chompakorn Chaksangchaichot}
\author[1]{Pirat Pothavorn}
\author[3]{Attapol Thamrongrattanarit-Rutherford}
\author[3]{Ekapol Chuangsuwanich}
\author[2]{Sarana Nutanong}

\affil[1]{VISAI AI, Thailand \authorcr
  \texttt{\{pawitsapaka\_visai,chompakornc\_pro,piratp\_visai\}@vistec.ac.th}}
\affil[2]{Vidyasirimedhi Institute of Science and Technology, Thailand \authorcr
  \texttt{\{sarana.n\}@vistec.ac.th}}
\affil[3]{Chulalongkorn University, Thailand \authorcr \texttt{\{attapol.t,ekapol.c\}@chula.ac.th}}

\title{Can Group Relative Policy Optimization Improve Thai Legal Reasoning and Question Answering?
}

\makeatletter
\renewcommand{\maketitle}{%
  \thispagestyle{fancy}%
  \vspace*{0.5em}
  \begin{center}
    \begin{tabular}{@{}m{1.8cm} m{0.75\textwidth}@{}}    
      \raggedright
      \includegraphics[width=1.5cm]{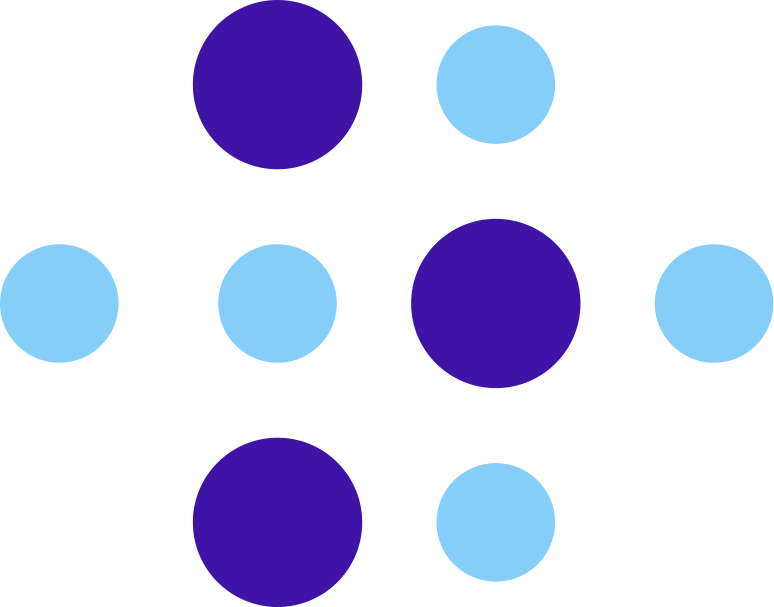} 
      & 
      {\bfseries \Large \@title}
      \\[4.0em]  

      \multicolumn{2}{@{}c@{}}{%
        \begin{minipage}{0.85\textwidth}
          \centering
          \@author \\[2.0em]
          \@date
        \end{minipage}
      } \\[1.0em]
    \end{tabular}
  \end{center}
  \vspace*{2em}
}
\makeatother

\begin{document}
\maketitle

\begin{abstract}
The Retrieval-Augmented Generation (RAG) systems' performance on Thai legal question answering is still limited, especially for questions requiring extensive, complex legal reasoning. 
To address these limitations, we introduce an approach aligning LLMs toward improved law citation accuracy and better response quality using Group-Relative Policy Optimization (GRPO). 
Our approach leverages BGE-M3 embeddings as a cost-efficient semantic-similarity reward, significantly reducing computational expenses up to 2.5x compared to large language model judges. 
Experiments on the NitiBench benchmark demonstrate substantial improvements: GRPO achieves up to 90\% citation-F1 gains from the base model and a 31\% increase in joint quality metrics over instruction tuning. 
Crucially, our method shows enhanced robustness on complex legal reasoning tasks compared to instruction tuning, providing an effective and resource-efficient solution for enhancing Thai legal LLMs.
\end{abstract}

\definecolor{darkgreen}{RGB}{10,100,10}
\definecolor{lightgray}{gray}{0.9}


\begin{figure*}[!hb]
    \centering
    \includegraphics[width=\textwidth]{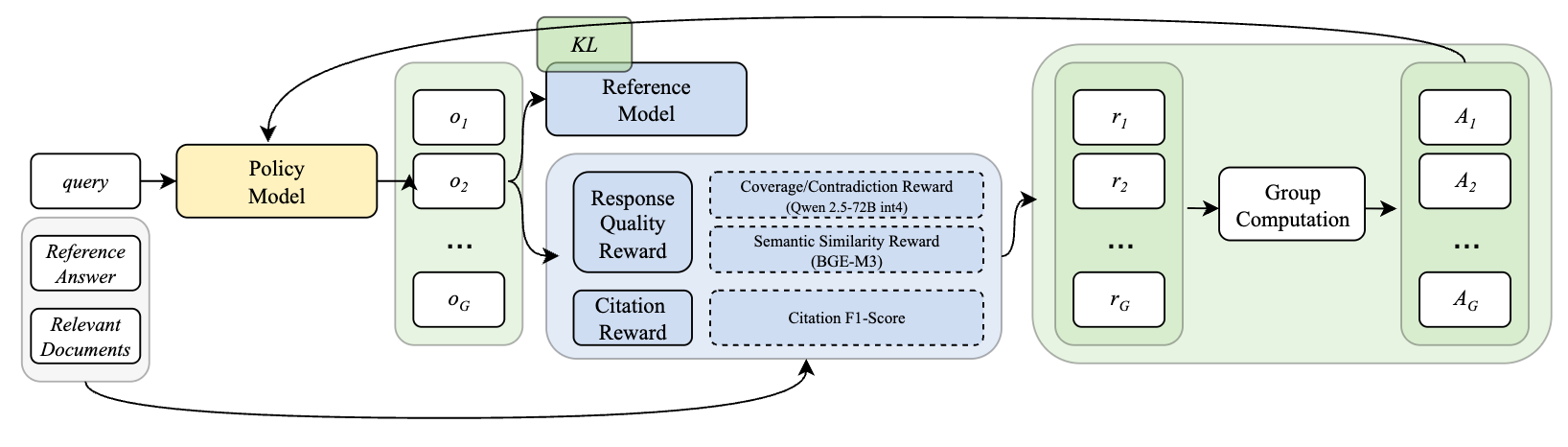} 
    \caption{Demonstration of our proposed method. Here, we use GRPO objectives with specialized reward to align LLM towards better citation and response using \textbf{Response Quality Reward} (\S\ref{subsec:rqr}) and \textbf{Citation Reward} (\S\ref{subsec:citation_reward}).}
    \label{fig:grpo}
\end{figure*}

\newpage
\tableofcontents
\newpage
\include{1-introduction}
\include{2-literature-review}
\include{3-methodology}
\include{4-experimental-setups}
\include{5-results}
\include{6-ablation-study}
\include{7-discussion}
\include{8-conclusion}
\include{9-limitation}

\bibliographystyle{plain}  
\bibliography{custom}

\include{10-appendix}

\end{document}

%% file: 1-introduction.tex
\section{Introduction}
\label{sec:introduction}

Recent advances in large language models (LLMs) have enabled new possibilities for legal question answering (QA) \cite{colombo2024saullm7bpioneeringlargelanguage,cap,HarveyAI}. 
However, delivering accurate and grounded responses remains challenging, especially in domains like Thai law, where legal complexity and limited training data lead to frequent hallucinations and citation errors \cite{nitibench}. 
Retrieval-augmented generation (RAG) \cite{rag} has been proposed to improve factuality, but existing systems often fail to cite relevant laws even when provided with the right context. 
While instruction tuning improves general fluency, it offers limited control over citation behavior. 
This motivates the need for more targeted alignment techniques that not only improve factuality but also enforce verifiable citation standards.

This discrepancy highlights the limited LLM performance in citing the correct documents necessary to answer the question.
Additionally, in many legal applications, the ability to correctly ground a response based on relevant law documents is critical, highlighting the need to improve LLM to improve its citation correctness.
With proper document citation, we suspect that this could potentially lead to better QA capability.


To address this, we consider whether reinforcement learning (RL) can provide more targeted control. 
Specifically, we explore Group-Relative Policy Optimization (GRPO) \cite{grpo}, which allows fine-grained reward shaping based on legal citation format, factual grounding, and answer quality~\cite{yasui-etal-2019-using}.

However, RL-based alignment often depends on high-quality reward signals, and in legal contexts, these typically come from expensive judge models.
This raises a second challenge: how to reduce the cost of alignment without sacrificing reward quality.
This work is driven by a central question: \emph{How can we affordably and effectively align large language models for citation-sensitive legal question answering, in a way that supports real-world deployment and domain-grounded evaluation in resource-constrained settings like Thai law?}


Based on the stated framing, we seek to answer the following research questions:
\begin{enumerate}[label=(RQ\arabic*)]
\item \textbf{GRPO vs Instruction Tuning:} How can reinforcement learning via GRPO improve citation accuracy and response quality in Thai legal question answering compared to instruction tuning?

\item \textbf{Thai-CPT vs Language-Agnostic:} To what extent does Thai-specific pretraining influence the effectiveness of GRPO and reward strategies in Thai legal QA?

\item \textbf{Reward Strategies:} What are the trade-offs between using a semantic similarity-based reward proxy and a large LLM-based reward model for aligning answer quality?

\end{enumerate}

\textbf{For RQ1}, we compare GRPO-based alignment with instruction tuning by applying a citation-oriented reward function, consisting of format adherence, hallucination checks, and citation F1, on top of instruction-tuned Thai legal LLMs. 
Evaluation is conducted on both in-domain and out-of-distribution benchmarks using the NitiBench benchmark \cite{nitibench}.
Results show that find that GRPO consistently improves citation fidelity and response quality, particularly on in-domain legal QA tasks.

\textbf{For RQ2}, we examine whether Thai-specific continued pretraining improves alignment outcomes.
All models share the Qwen2.5-7B architecture, and we compare a language-agnostic model to two Thai-CPT variants: Typhoon2 \cite{typhoon2} and OpenThaiGPT1.5 \cite{openthaigpt15}.
We find that the Thai-CPT models outperform the language-agnostic baseline under both reward strategies, indicating greater receptiveness to alignment signals.

\textbf{For RQ3}, we evaluate two GRPO reward strategies: a cost-efficient semantic similarity proxy (BGE-M3) and a judge-based reward from Qwen2.5-72B-Instruct that scores coverage and contradiction. 
This comparison allows us to assess alignment effectiveness relative to training cost.
We observe that the semantic proxy achieves comparable alignment quality with significantly lower GPU cost.

In summary, our contributions are as follows.

\noindent
\textbf{Problem Formulation:} We frame Thai legal QA as a citation-sensitive alignment problem and propose using GRPO to guide model behavior through verifiable reward signals, enabling effective and affordable supervision in low-resource legal settings.

\noindent
\textbf{Problem Decomposition:} Based on this framing, we investigate the impact of GRPO over instruction tuning, the role of Thai-specific pretraining in shaping alignment effectiveness, and the trade-offs between semantic and judge-based reward signals.

\noindent
\textbf{Key Insights:} GRPO consistently improves citation fidelity and response quality; and Thai-CPT models are more receptive to alignment signals than their language-agnostic counterparts; semantic rewards offer comparable alignment to judge models with significantly lower cost.
\emph{These results underscore the importance of both targeted alignment and domain-specific pretraining in building robust legal QA RAG systems in a resource-constrained language like Thai.}


%

%% file: 2-literature-review.tex
\section{Related Work}
\label{sec:related_work}


Large Language Models (LLMs), typically based on the Transformer architecture \cite{attention}, have advanced NLP but face challenges in specialized domains like Thai Legal Question Answering (QA) due to data scarcity and the need for complex, grounded reasoning \cite{nitibench}. 
While Retrieval-Augmented Generation (RAG) \cite{rag} aims to improve grounding by retrieving external knowledge, and Long-Context Language Models (LCLMs) \cite{gemini15} handle longer inputs, limitations in consistent reasoning and reliability in the legal domain \cite{nitibench}. 
The Nitibench \cite{nitibench}, a benchmark focusing on Thai law with datasets like Nitibench-CCL and Nitibench-Tax, provides essential tools for evaluating models on tasks requiring nuanced legal understanding and accurate source attribution, highlighting the difficulties compared to benchmarks in resource-rich languages like English \cite{lexglue, legalbench}.

\paragraph{Enhancing LLM legal citation performance.}
A growing body of work seeks to make LLMs produce verifiable citations.
CitaLaw \cite{zhang2025citalawenhancingllmcitations} adapts the ALCE benchmark \cite{gao-etal-2023-enabling} to the legal domain, introduces a syllogism-level citation metric, and supports both statutes and precedent cases.
ALCE itself evaluates statement-level grounding using an NLI verifier, requiring every generated claim to be backed by retrieved evidence.
\cite{shareghi2024methodslegalcitationprediction} compare citation accuracy across three retrieval regimes: 1) retriever-only, 2) LLM query-rewrite, and 3) hybrid method.
This work focus on Australian case-law, and show that task-specific instruction tuning yields the largest gains in improving citation accuracy.
LegalBench-RAG \cite{pipitone2024legalbenchragbenchmarkretrievalaugmentedgeneration} isolates the retriever’s contribution by measuring precision over expert-annotated snippets while varying chunking and top-k, revealing a retrieval-quality ceiling on downstream citation F1.


\paragraph{Usage of embedding-based reward models.}
Early works explored leveraging pretrained embeddings as reward signals for text generation alignment.
\cite{yasui-etal-2019-using} finetune BERT \cite{devlin2019bertpretrainingdeepbidirectional} on Semantic Textual Similarity (STS) and employ the tuned model as a REINFORCE reward for machine translation.
\cite{cs2292019bert} optimize an abstractive summarizer directly for BERTScore \citep{zhang2020bertscoreevaluatingtextgeneration}, observing higher fluency and lower redundancy than ROUGE-reward baselines.
More recently, \cite{sun2025reusingembeddingsreproduciblereward} distil preference scores from the ``gold’’ reward model of \cite{dong2023raftrewardrankedfinetuning,dong2024rlhfworkflowrewardmodeling} into lightweight proxies, an MLP and a LightGBM, that take paired Gemma-2B embeddings as input, achieving judge-level quality.
These results indicate that inexpensive embedding-based rewards can rival heavyweight LLM judges for preference-optimized generation.
However, their integration into modern preference-optimization algorithms remains under-explored.

\paragraph{Research Gap.} Existing legal QA systems rely primarily on instruction tuning, with limited success on citation grounding. 
While GRPO and efficient reward proxies have shown promise elsewhere, their application to legal domains, particularly Thai legal QA, remains underexplored. 
We address this gap by investigating GRPO’s impact on citation accuracy and evaluating cost-effective reward strategies.

%% file: 3-methodology.tex
\section{Proposed Studies}
\label{sec:methodology}

%
%
%

We frame the Thai legal question answering (QA) task as a citation-sensitive generative task, where the model must generate a free-form response based on a user query and a set of retrieved legal documents. 
The response must be semantically informative, cite relevant statutes, and avoid hallucinations that reference unsupported claims.
To align LLMs with these objectives, we proposed a framework based on GRPO, which aligns model behavior through a carefully designed reward function that encourages citation accuracy and response quality.
Our proposed method is summarized in Figure \ref{fig:grpo}.

\label{sec:grpo_rewards}

%

Formally, given a dataset $\mathcal{D}$, a reward function $R(s,a)$ where:
\begin{itemize}
    \item $s$: Current state, a combination of LLM prompt, user query/question, and relevant document obtained from the embeddings model.
    \item $a$: Next action, the generated response to the question, and citation to the generated question.
\end{itemize}
We aim to learn an optimal policy $\pi$ that maximizes the reward function $R(s,a)$  using GRPO objectives \cite{grpo}.
This reward-based formulation enables us to study how different alignment strategies influence legal response quality under the constraints of low-resource legal domains. The next sections describe our design choices and experimental findings.


\subsection{Citation Accuracy Reward (RQ1)}
\label{subsec:citation_reward}

Let us examine whether fine-grained reward shaping allows the model to receive more targeted feedback, especially in areas where factual correctness and citation structure are essential.
To that end, we design a multi-component reward function that captures core requirements for verifiable legal citation.

The design is motivated by the need to overcome limitations of instruction tuning, which can sometimes lead to poor generalization \cite{kalajdzievski2024scalinglawsforgettingfinetuning}, especially in out-of-distribution legal reasoning tasks \cite{liu-etal-2024-catastrophic}.
In particular, our reward formulation decomposes citation quality into three measurable dimensions:
\begin{itemize}
    
    \item \textbf{Format Reward} [0.0, 1.0]: Assesses adherence to the required XML structure. This is a prerequisite for subsequent citation checks.
    \item \textbf{Non-Hallucination Reward} [0.0, 0.5]: Contingent on passing Format check, this reward checks whether the cited law was actually provided in the context or not. In other words, we penalize any generated citations that are not included in the retrieved law section augmented to the prompt. 
    \item \textbf{Citation F1 Reward} [0.0 - 1.0]: Contingent on passing both previous checks, this measures the F1 score between cited law sections and the ground truth.
\end{itemize}
To assess the effectiveness, we train three Thai legal LLMs using GRPO guided by this reward function and compare them against instruction-tuned baselines (SFT and PEFT) on both in-domain (NitiBench-CCL) and out-of-distribution (NitiBench-Tax) benchmarks.

\subsection{Thai-Specific vs. Language-Agnostic (RQ2)}
\label{sec:model_selection}

For RQ2, we evaluate two types of pretrained models: (i) a language-agnostic model directly instruction-tuned from Qwen2.5-7B, and (ii) Thai-specific models that undergo continued pretraining on Thai-centric corpora prior to instruction tuning.
\begin{enumerate}
    \item Qwen2.5-7B-Instruct\footnote{https://huggingface.co/Qwen/Qwen2.5-7B-Instruct} \cite{qwen25}
    \item Typhoon2-qwen2.5-7b-Instruct\footnote{https://huggingface.co/scb10x/typhoon2-qwen2.5-7b-instruct} \cite{typhoon2}
    \item OpenThaiGPT1.5-7b-Instruct\footnote{https://huggingface.co/openthaigpt/openthaigpt1.5-7b-instruct} \cite{openthaigpt15}
\end{enumerate}

All three models are based on the Qwen2.5-7B architecture to ensure architectural consistency.  
The Thai-specific models—Typhoon2 and OpenThaiGPT1.5—allow us to examine whether language-specific continued pretraining influences the effectiveness of GRPO-based alignment and reward strategies in Thai legal QA. 
Since all models share the same architecture and are trained under identical alignment and evaluation settings, this setup isolates pretraining data as the key factor under investigation for RQ3, while maintaining full comparability with the studies addressing RQ1 and RQ2.

\subsection{Response Quality Reward (RQ3)}
\label{subsec:rqr}

In addition to the citation accuracy reward, we also design a reward to ensure that the generated response quality is acceptable given the reference answer from the ground truth.
While ideal judges like preference models or advanced reasoning LLMs (e.g., OpenAI o1 \cite{o1}, Deepseek R1 \cite{deepseekr1}) are too slow or costly for online training, we explore more computationally efficient proxies.
Instead of using an LLM judge to score rewards, which can be expensive, we proposed to use the semantic similarity between roll-out responses and the reference answer from the ground truth as a reward instead.
\begin{itemize}
    \item \textbf{Semantic Similarity Reward} [0.0 - 1.0]: As one proxy, we use the BGE-M3 \cite{bge-m3} embedding model (with a multi-head strategy) to calculate the similarity score between the generated answer text and the ground-truth answer.
\end{itemize}
We call this setup of using embeddings as a reward model a ``semantic reward."

Alternatively, we also investigate the use of coverage and contradiction scores (following \cite{nitibench}) directly as rewards. 
To make this feasible during training, we employ Qwen2.5-72B-Instruct\footnote{Deployed on A100-80GB GPU with int4\_wo precision with TensorRT-LLM \cite{tensorrtllm}.} \cite{qwen25} as an approximation to the strong LLM judge.
Using LLM judge, we directly optimize for the answer coverages and contradiction following Nitibench metrics \cite{nitibench}.
\begin{itemize}
    \item \textbf{Coverage Reward} [0.0, 0.5, 1.0]: Measures semantic coverage between the generated and ground truth answer, whether the generated response is \textit{not coverage}, \textit{partially coverage}, or \textit{fully coverage}.
    \item \textbf{Contradiction Reward} [0.0, 1.0]: A binary reward assesses whether the generated response contradicts the ground truth answer or not.
\end{itemize}
This setup of using LLM as a reward model is named ``cov/con reward.''

%% file: 4-experimental-setups.tex
\section{Experimental Setup}
\label{sec:experimental_setup}

This section outlines the configuration for training and evaluating models using Instruction Tuning (IT) and Group Relative Policy Optimization (GRPO).

\subsection{Training Data and Benchmark}
\label{sec:data}
We use WangchanX-Legal-ThaiCCL-RAG \cite{nitibench} as a training set for both IT and GRPO setups.
Data entries contain a question, a ground-truth relevant legal section, and a reference answer. 
When preparing the training set, we construct prompts using the question and top-retrieved sections instead of the ground-truth law sections.
We use BGE-M3 \cite{bge-m3} with a multi-head strategy (dense/sparse/ColBERT weights set to 0.4, 0.2, 0.4) to retrieve the top 10 relevant law sections.
The ground-truth sections are used for citation evaluation/reward, and the reference answer is used for answer evaluation/reward. 

For the benchmark, we utilize \textbf{Nitibench}\footnote{https://huggingface.co/datasets/VISAI-AI/nitibench} dataset \cite{nitibench}, specifically designed for Thai Legal QA. 
The benchmark contains two splits:
\begin{itemize}
    \item \textbf{Nitibench-CCL:} Focuses on general Thai corporate/commercial law.
    \item \textbf{Nitibench-Tax:} Comprises complex, multi-positive Thai tax rulings. Used exclusively as a test set to evaluate generalization to a very complex legal reasoning task.
\end{itemize}

%

\subsection{Evaluation Metrics}
\label{sec:evaluation_metrics}
We adopt the End-to-End (E2E) metrics from Nitibench \cite{nitibench}.
However, instead of using a contradiction score, we use \textit{consistency score}, which is an inverse of the contradiction score ($\text{Consistency Score}=1-\text{Contradiction Score}$).
The reason behind this is to normalize all E2E metrics to be within the range of 0 to 1, which allows us to further calculate \textit{Joint Score}, which is an average of Citation F1, Coverage score, and Consistency score.
Each metric is described as follows.
\begin{itemize}
    \item \textbf{Citation F1:} F1-score of cited legal sections compared to the ground truth.
    \item \textbf{Coverage:} Reference answer overlap between generated and ground-truth answers based on a 0/50/100 scale. The value was then normalized to range from 0 to 1.
    \item \textbf{Consistency:} Factual consistency of the generated answer with the ground truth. Calculated as \texttt{1 - Contradiction}, leveraging the Contradiction score from Nitibench, where 0: No-Contradiction, 1: Contradiction.
    \item \textbf{Joint Score:} An average of three metrics above.
\end{itemize}
We are also employing GPT-4o as the judge with Nitibench prompts, ensuring consistency and comparability with the original benchmark. 
These metrics assess crucial aspects of legal answer quality: grounding (Citation F1), answer correctness (Coverage), and factual reliability (Consistency).

\subsection{Training Setups}

\paragraph{Prompt Construction}
To manage computational constraints, input prompts are capped at 8192 tokens. 
If the retrieved top 10 sections exceed this limit, we iteratively replace the longest nonground-truth section with the next highest-ranked section from the retriever, ensuring all ground-truth sections are retained while staying within the token limit.
The target output format for both IT and GRPO is structured XML-like text including \texttt{<reasoning>}, \texttt{<answer>}, and \texttt{<citation>} tags. 
Additional details regarding input and output formatting are provided in Appendix \ref{appendix:input_output_format}.

\paragraph{Training Objectives}
We use Parameter-Efficient Fine-Tuning (PEFT) via Low-Rank Adaptation (LoRA) \cite{lora} (rank \(r=256\), applied to attention layers\footnote{We apply LoRA on \texttt{q\_proj}, \texttt{k\_proj}, \texttt{v\_proj}, \texttt{gate}, \texttt{up\_proj}, \texttt{down\_proj}} with 16-bit precision). 
This significantly reduces memory for GRPO as the policy model will update on the adapter instead of full model weights. 
All GRPO setups was trained using Unsloth \cite{unsloth} on a single NVIDIA A100 80gb GPUs.
We trained GRPO for one epoch using a learning rate of 5e-6, a max gradient normalization of 0.2, a \texttt{`constant\_with\_warmup'} learning rate scheduler, and a rollout amount of 10.
All hyperparameters can be found in Appendix \ref{appendix:training_params}. 

\paragraph{Baseline}
For the baseline, we used 
\begin{itemize}
    \item \textbf{Base Instruction Tuned Model:} The base instructioned tuned model provided by original authors.
    \item \textbf{Instruction Tuned Model with LoRA:} The instruction tuned model with LoRA adapter targeting the same modules with the same rank configuration ($r=256$). We finetuned for 3 epochs on the training set.
\end{itemize}

\subsection{Inference and Result Aggregation}
\label{sec:inference_aggregation}
To ensure robust evaluation, we performed inference 3 times for each model configuration on both the Nitibench-CCL and Nitibench-Tax test sets. Each run utilized a different generation random seed (see Appendix \ref{appendix:inferencing_params} for details) and was executed using vLLM \cite{vllm}. 
We report the final performance as the mean and standard deviation of the scores across these three runs.

%% file: 5-results.tex
\input{table/main_table}
\section{Results}
\label{sec:results}

This section presents a comparison between the baseline performance and our proposed method. 
Detailed results, including mean and standard deviation across 3 runs, are provided in Table~\ref{tab:main_table}.

\underline{Note that} the Citation F1 metric is inherently limited by the performance of the upstream BGE-M3 retriever, which achieves an F1 score of 0.9220 on Nitibench-CCL and 0.4809 on Nitibench-Tax.
This represents the theoretical upper bound for Citation F1 that the LLM could achieve, as it cannot cite documents not provided by the retriever.

\subsection{Effectiveness of GRPO}

\paragraph{GRPO improves the performance of the LLM more compared to instruction tuning under in-domain setups.}
%
%
As shown in Table~\ref{tab:main_table}, on the in-domain \textbf{Nitibench-CCL} test set, both GRPO variants consistently achieve substantial improvements over instruction tuning (e.g., +81-90\% gain for Typhoon2 GRPO vs. +60\% for instruction tuned on Citation F1) and also significantly boost Coverage and Consistency, resulting in much higher Joint Scores compared to instruction tuning.

\paragraph{GRPO shows a mixed but consistent improvement over instruction tuning on NitiBench-Tax}
Table~\ref{tab:main_table} also shows that while instruction tuning consistently degrades performance across all metrics compared to the baseline (large negative gains, e.g., -54\% Citation F1 for Qwen2.5-7B-Instruct), GRPO shows a markedly different behavior.

\paragraph{GRPO-tuned 7B model performance almost comparable to the best proprietary model.}
Comparing against results for large proprietary models (GPT-4o, Gemini 1.5 Pro, Claude 3.5 Sonnet)\footnote{Results were taken from NitiBench paper \cite{nitibench}.}, our fine-tuned 7B models with GRPO achieve highly competitive scores on \textbf{Nitibench-CCL}, sometimes approaching or exceeding models like Claude 3.5 Sonnet and nearing Gemini 1.5 Pro as shown in Table~\ref{tab:main_table}.

\paragraph{Our method still can't generalize well compared to proprietary models on a more challenging NitiBench-Tax}
On the challenging \textbf{Nitibench-Tax} generalization task, all listed larger models significantly outperform our tuned 7B models, highlighting the difficulty of this task and indicating significant room for future improvement.




\subsection{Impact of Base Model Priors}

\paragraph{GRPO Enhances Sampling Efficiency for Thai-Aligned Base Models on Nitibench-Tax.}
Table~\ref{tab:main_table} offers a comparison between a language-agnostic model (Qwen2.5) and Thai-aligned models (Typhoon2, OpenThaiGPT1.5).
In particular, the Nitibench-Tax results highlight GRPO's effectiveness depending on base model capabilities, aligning with findings suggesting RL enhances sampling efficiency rather than reasoning capacity \cite{Yue2025LimitRLVR}. 
While GRPO struggled with Qwen2.5 on Nitibench-Tax, it yielded significant gains over baseline on the Thai-aligned models.

This suggests aligned models possess a stronger inherent capacity (better priors) for these complex tasks. 
GRPO appears to primarily enhance sampling efficiency, biasing output towards existing correct pathways more frequently, thus improving observed performance (like Citation F1) for problems likely solvable within the base model's capacity. 
In contrast, instruction tuning consistently failed, potentially disrupting paths or overfitting. 
However, GRPO's modest Coverage/Consistency gains on Tax suggest improved sampling alone doesn't fully address complex answer quality, hinting at RL's potential limitations in expanding the reasoning boundary as noted by \cite{Yue2025LimitRLVR}.

\input{table/ablation_table}

\subsection{GRPO Reward Strategies}

\paragraph{Semantic Similarity Reward shows a comparable performance with Coverage + Consistency reward, especially in in-domain.} 
Let us now focus on the two GRPO variants \emph{coverage + consistency} and \emph{semantic similarity} in Table~\ref{tab:main_table}.
The GRPO variant using \textbf{semantic similarity reward} often achieves the highest scores on the \textbf{Nitibench-CCL} set, particularly boosting Coverage significantly (e.g., +38\% gain for Typhoon2). 
This highlights the potential of using efficient semantic similarity as a reward proxy for in-domain tasks where the ground-truth answer provides a strong semantic target, offering a cost-effective alternative to employing large judge models like Qwen2.5-72B-Instruct during training.

\paragraph{Coverage + Consistency reward yield sturdier generalization.}
On the more challenging \textbf{Nitibench-Tax} set, the results are more mixed. 
While the semantic reward variant sometimes leads in Citation F1 for aligned models (e.g., OpenThaiGPT1.5), the GRPO variant using \textbf{coverage and consistency rewards} from the Qwen-72B judge occasionally shows slightly better robustness in maintaining Coverage and Consistency scores under generalization pressure. 
This suggests that while semantic similarity is a promising low-cost signal, directly optimizing for Coverage and Consistency using a capable (albeit more expensive) judge model might offer advantages for complex generalization tasks, though both GRPO strategies vastly outperform instruction-tuned models.

%% file: table/main_table.tex
\begin{table*}[!ht]
\centering
\resizebox{\textwidth}{!}{%
\begin{tabular}{@{}lccccccccccc@{}}
\toprule
\rowcolor[HTML]{FFFFFF} 
\textbf{model} &
  Citation F1 $\uparrow$ &
  SD &
  gains (\%) &
  Coverage $\uparrow$ &
  SD &
  gains (\%) &
  Consistency $\uparrow$ &
  SD &
  gains (\%) &
  Joint score &
  gains (\%) \\ \midrule
\rowcolor[HTML]{FFFFFF} 
\multicolumn{12}{c}{\cellcolor[HTML]{FFFFFF}\textbf{Nitibench-CCL (In-Domain)}} \\ \midrule
\rowcolor[HTML]{D9D9D9} 
{\color[HTML]{000000} qwen2.5-7b-instruct} &
  {\color[HTML]{000000} 0.4103} &
  {\color[HTML]{000000} 0.0015} &
  {\color[HTML]{FFFFFF} } &
  {\color[HTML]{000000} 0.5908} &
  {\color[HTML]{000000} 0.0041} &
  {\color[HTML]{000000} } &
  {\color[HTML]{000000} 0.8402} &
  0.0030 &
  {\color[HTML]{00B050} \textbf{}} &
  0.6138 &
   \\
\rowcolor[HTML]{FFFFFF} 
+LoRA SFT &
  0.5691 &
  0.0040 &
  {\color[HTML]{4EA72E} 38.70} &
  0.5832 &
  0.0075 &
  {\color[HTML]{FF0000} -1.29} &
  0.8341 &
  0.0024 &
  {\color[HTML]{FF0000} -0.72} &
  0.6622 &
  {\color[HTML]{4EA72E} 7.88} \\
\rowcolor[HTML]{FFFFFF} 
+LoRA GRPO (cov/con reward) &
  0.6796 &
  0.0020 &
  {\color[HTML]{4EA72E} 65.63} &
  0.6322 &
  0.0010 &
  {\color[HTML]{4EA72E} 7.00} &
  0.8598 &
  0.0009 &
  {\color[HTML]{4EA72E} 2.34} &
  0.7239 &
  {\color[HTML]{4EA72E} 17.94} \\
\rowcolor[HTML]{FFFFFF} 
+LoRA GRPO (semantic reward) &
  {\ul 0.7146} &
  0.0009 &
  {\color[HTML]{4EA72E} 74.14} &
  0.7197 &
  0.0023 &
  {\color[HTML]{00B050} 21.81} &
  0.8232 &
  0.0024 &
  {\color[HTML]{FF0000} -2.02} &
  0.7525 &
  {\color[HTML]{4EA72E} 22.60} \\
\rowcolor[HTML]{D9D9D9} 
typhoon2-qwen2.5-7b-instruct &
  0.3597 &
  0.0042 &
  {\color[HTML]{4EA72E} } &
  0.5587 &
  0.0061 &
  {\color[HTML]{4EA72E} } &
  0.8553 &
  0.0076 &
  \cellcolor[HTML]{D9D9D9}{\color[HTML]{00B050} } &
  0.5912 &
  {\color[HTML]{4EA72E} } \\
\rowcolor[HTML]{FFFFFF} 
+LoRA SFT &
  0.5744 &
  0.0028 &
  {\color[HTML]{4EA72E} 59.71} &
  0.6214 &
  0.0030 &
  {\color[HTML]{4EA72E} 11.23} &
  0.8572 &
  0.0030 &
  {\color[HTML]{4EA72E} 0.22} &
  0.6843 &
  {\color[HTML]{4EA72E} 15.75} \\
\rowcolor[HTML]{FFFFFF} 
+LoRA GRPO (cov/con reward) &
  0.6514 &
  0.0013 &
  {\color[HTML]{4EA72E} 81.10} &
  0.7092 &
  0.0039 &
  {\color[HTML]{4EA72E} 26.95} &
  \textbf{0.9032} &
  0.0019 &
  {\color[HTML]{4EA72E} 5.60} &
  0.7546 &
  {\color[HTML]{4EA72E} 27.63} \\
\rowcolor[HTML]{FFFFFF} 
+LoRA GRPO (semantic reward) &
  0.6828 &
  0.0028 &
  {\color[HTML]{4EA72E} 89.84} &
  \textbf{0.7735} &
  0.0012 &
  {\color[HTML]{4EA72E} 38.45} &
  {\ul 0.8757} &
  0.0028 &
  {\color[HTML]{4EA72E} 2.38} &
  \textbf{0.7773} &
  {\color[HTML]{4EA72E} 31.48} \\
\rowcolor[HTML]{D9D9D9} 
openthaigpt1.5-qwen2.5-7b-instruct &
  0.4299 &
  0.0048 &
  {\color[HTML]{4EA72E} \textbf{}} &
  0.5556 &
  0.0010 &
  {\color[HTML]{00B050} \textbf{}} &
  0.8234 &
  0.0048 &
  \cellcolor[HTML]{D9D9D9}{\color[HTML]{4EA72E} \textbf{}} &
  0.6030 &
  {\color[HTML]{4EA72E} } \\
\rowcolor[HTML]{FFFFFF} 
{\color[HTML]{000000} +LoRA SFT} &
  {\color[HTML]{000000} 0.5613} &
  {\color[HTML]{000000} 0.0069} &
  {\color[HTML]{4EA72E} 30.56} &
  {\color[HTML]{000000} 0.5930} &
  {\color[HTML]{000000} 0.0024} &
  {\color[HTML]{4EA72E} 6.73} &
  {\color[HTML]{000000} 0.8371} &
  {\color[HTML]{000000} 0.0031} &
  {\color[HTML]{4EA72E} 1.66} &
  0.6638 &
  {\color[HTML]{4EA72E} 10.08} \\
\rowcolor[HTML]{FFFFFF} 
+LoRA GRPO (cov/con reward) &
  \textbf{0.7197} &
  0.0020 &
  {\color[HTML]{4EA72E} 67.40} &
  0.6680 &
  0.0034 &
  {\color[HTML]{4EA72E} 20.23} &
  0.8705 &
  0.0034 &
  {\color[HTML]{4EA72E} 5.72} &
  0.7527 &
  {\color[HTML]{4EA72E} 24.84} \\
\rowcolor[HTML]{FFFFFF} 
+LoRA GRPO (semantic reward) &
  0.7017 &
  0.0016 &
  {\color[HTML]{4EA72E} 63.23} &
  {\ul 0.7214} &
  0.0041 &
  {\color[HTML]{4EA72E} 29.84} &
  0.8554 &
  0.0021 &
  {\color[HTML]{4EA72E} 3.89} &
  {\ul 0.7595} &
  {\color[HTML]{4EA72E} 25.96} \\ \midrule \midrule
\rowcolor[HTML]{FFFFFF} 
gpt-4o-2024-08-06 &
  0.7140 &
   &
   &
  0.8520 &
   &
   &
  0.9450 &
  \textbf{} &
  \textbf{} &
  0.8370 &
  \textbf{} \\
\rowcolor[HTML]{FFFFFF} 
gemini-1.5-pro-002 &
  0.6510 &
   &
   &
  0.8650 &
   &
   &
  0.9520 &
  \textbf{} &
  \textbf{} &
  0.8227 &
  \textbf{} \\
\rowcolor[HTML]{FFFFFF} 
claude-3-5-sonnet-20240620 &
  0.5950 &
   &
   &
  0.8970 &
   &
   &
  0.9600 &
  \textbf{} &
  \textbf{} &
  0.8173 &
  \textbf{} \\ \midrule
\rowcolor[HTML]{FFFFFF} 
\multicolumn{12}{c}{\cellcolor[HTML]{FFFFFF}\textbf{Nitibench-Tax (Out-of-Domain)}} \\ \midrule
\rowcolor[HTML]{D9D9D9} 
qwen2.5-7b-instruct &
  0.2110 &
  0.0272 &
  {\color[HTML]{FF0000} } &
  0.3333 &
  0.0082 &
  {\color[HTML]{00B050} \textbf{}} &
  0.5733 &
  0.0340 &
  {\color[HTML]{00B050} } &
  0.3726 &
   \\
\rowcolor[HTML]{FFFFFF} 
+LoRA SFT &
  0.0975 &
  0.0192 &
  {\color[HTML]{FF0000} -53.82} &
  0.2867 &
  0.0249 &
  {\color[HTML]{FF0000} -13.99} &
  0.5067 &
  0.0094 &
  {\color[HTML]{FF0000} -11.63} &
  0.2969 &
  {\color[HTML]{FF0000} -20.30} \\
\rowcolor[HTML]{FFFFFF} 
+LoRA GRPO (cov/con reward) &
  0.1678 &
  0.0196 &
  {\color[HTML]{FF0000} -20.47} &
  0.2933 &
  0.0047 &
  {\color[HTML]{FF0000} -12.00} &
  0.5633 &
  0.0094 &
  {\color[HTML]{FF0000} -1.74} &
  0.3415 &
  {\color[HTML]{FF0000} -8.34} \\
\rowcolor[HTML]{FFFFFF} 
+LoRA GRPO (semantic reward) &
  0.1555 &
  0.0135 &
  {\color[HTML]{FF0000} -26.31} &
  0.3167 &
  0.0249 &
  {\color[HTML]{FF0000} -4.99} &
  0.5667 &
  0.0249 &
  {\color[HTML]{FF0000} -1.16} &
  0.3463 &
  {\color[HTML]{FF0000} -7.05} \\
\rowcolor[HTML]{D9D9D9} 
typhoon2-qwen2.5-7b-instruct &
  0.1272 &
  0.0150 &
  {\color[HTML]{4EA72E} \textbf{}} &
  0.3333 &
  0.0411 &
  {\color[HTML]{FF0000} } &
  0.5467 &
  0.0249 &
  \cellcolor[HTML]{D9D9D9}{\color[HTML]{4EA72E} } &
  0.3357 &
   \\
\rowcolor[HTML]{FFFFFF} 
+LoRA SFT &
  {\color[HTML]{000000} 0.1072} &
  {\color[HTML]{000000} 0.0315} &
  {\color[HTML]{FF0000} -15.71} &
  {\color[HTML]{000000} 0.2633} &
  {\color[HTML]{000000} 0.0205} &
  {\color[HTML]{FF0000} -21.00} &
  {\color[HTML]{000000} 0.5667} &
  {\color[HTML]{000000} 0.0189} &
  {\color[HTML]{4EA72E} 3.66} &
  0.3124 &
  {\color[HTML]{FF0000} -6.95} \\
\rowcolor[HTML]{FFFFFF} 
+LoRA GRPO (cov/con reward) &
  0.2035 &
  0.0197 &
  {\color[HTML]{4EA72E} 60.03} &
  \textbf{0.3800} &
  0.0294 &
  {\color[HTML]{00B050} 14.00} &
  {\ul 0.5833} &
  0.0189 &
  {\color[HTML]{4EA72E} 6.71} &
  \textbf{0.3889} &
  {\color[HTML]{4EA72E} 15.85} \\
\rowcolor[HTML]{FFFFFF} 
+LoRA GRPO (semantic reward) &
  {\ul 0.2113} &
  0.0134 &
  {\color[HTML]{4EA72E} 66.18} &
  0.3633 &
  0.0411 &
  {\color[HTML]{4EA72E} 9.00} &
  0.4933 &
  0.0525 &
  {\color[HTML]{FF0000} -9.76} &
  0.3560 &
  {\color[HTML]{4EA72E} 6.04} \\
\rowcolor[HTML]{D9D9D9} 
openthaigpt1.5-qwen2.5-7b-instruct &
  0.1850 &
  0.0247 &
   &
  0.3367 &
  0.0519 &
   &
  0.5400 &
  0.0849 &
  \cellcolor[HTML]{D9D9D9} &
  0.3539 &
   \\
\rowcolor[HTML]{FFFFFF} 
+LoRA SFT &
  0.1039 &
  0.0387 &
  {\color[HTML]{FF0000} -43.84} &
  0.3267 &
  0.0450 &
  {\color[HTML]{FF0000} -2.97} &
  0.5800 &
  0.0283 &
  {\color[HTML]{4EA72E} 7.41} &
  0.3368 &
  {\color[HTML]{FF0000} -4.81} \\
\rowcolor[HTML]{FFFFFF} 
+LoRA GRPO (cov/con reward) &
  0.2085 &
  0.0328 &
  {\color[HTML]{4EA72E} 12.73} &
  {\ul 0.3667} &
  0.0205 &
  {\color[HTML]{00B050} 12.24} &
  0.5600 &
  0.0748 &
  {\color[HTML]{4EA72E} 3.70} &
  {\ul 0.3784} &
  {\color[HTML]{4EA72E} 6.93} \\
\rowcolor[HTML]{FFFFFF} 
+LoRA GRPO (semantic reward) &
  \textbf{0.2482} &
  0.0054 &
  {\color[HTML]{4EA72E} 34.16} &
  0.2500 &
  0.0424 &
  {\color[HTML]{FF0000} -25.74} &
  \textbf{0.6000} &
  0.0490 &
  {\color[HTML]{4EA72E} 11.11} &
  0.3661 &
  {\color[HTML]{4EA72E} 3.44} \\ \midrule \midrule
\rowcolor[HTML]{FFFFFF} 
gpt-4o-2024-08-06 &
  0.4380 &
   &
   &
  0.5000 &
   &
   &
  0.5400 &
  \textbf{} &
  \textbf{} &
  0.4927 &
  \textbf{} \\
\rowcolor[HTML]{FFFFFF} 
gemini-1.5-pro-002 &
  0.3320 &
   &
  {\color[HTML]{FF0000} } &
  0.4400 &
   &
  {\color[HTML]{FF0000} } &
  0.5200 &
  \textbf{} &
  {\color[HTML]{00B050} \textbf{}} &
  0.4307 &
  \textbf{} \\
\rowcolor[HTML]{FFFFFF} 
claude-3-5-sonnet-20240620 &
  0.4570 &
   &
  {\color[HTML]{FF0000} \textbf{}} &
  0.5100 &
   &
  {\color[HTML]{FF0000} \textbf{}} &
  0.5600 &
  \textbf{} &
  {\color[HTML]{00B050} \textbf{}} &
  0.5090 &
  \textbf{} \\ \bottomrule
\end{tabular}%
}
\caption{Performance comparison (avg 3 runs) on Nitibench-CCL and Nitibench-Tax: Baseline vs. SFT, GRPO (cov/con reward), GRPO (semantic reward). Relative performance gains over baseline are indicated. Comparison provided against 3 proprietary LLM results from \cite{nitibench} on the same settings.}
\label{tab:main_table}
\end{table*}

%% file: table/ablation_table.tex
\begin{table*}[!ht]
\centering
\resizebox{\textwidth}{!}{%
\begin{tabular}{@{}lccccccccccc@{}}
\toprule
\rowcolor[HTML]{FFFFFF} 
model &
  Citation F1 $\uparrow$ &
  SD &
  gains (\%) &
  Coverage $\uparrow$ &
  SD &
  gains (\%) &
  Consistency $\uparrow$ &
  SD &
  gains (\%) &
  Joint score &
  gains (\%) \\ \midrule
\rowcolor[HTML]{FFFFFF} 
\multicolumn{12}{c}{\cellcolor[HTML]{FFFFFF}\textbf{Nitibench-CCL (In-Domain)}} \\ \midrule
\rowcolor[HTML]{D9D9D9} 
openthaigpt1.5-qwen2.5-7b-instruct &
  0.4299 &
  0.0048 &
   &
  0.5556 &
  0.0010 &
   &
  0.8234 &
  0.0048 &
   &
  0.6030 &
   \\
\rowcolor[HTML]{EFEFEF} 
+LoRA SFT &
  0.5613 &
  0.0069 &
  {\color[HTML]{4EA72E} 30.56} &
  0.5930 &
  0.0024 &
  {\color[HTML]{4EA72E} 6.73} &
  0.8371 &
  0.0031 &
  \cellcolor[HTML]{F2F2F2}{\color[HTML]{4EA72E} 1.66} &
  0.6638 &
  {\color[HTML]{4EA72E} 10.08} \\
\rowcolor[HTML]{EFEFEF} 
+LoRA GRPO (cov/con reward) &
  \textbf{0.7197} &
  0.0020 &
  {\color[HTML]{4EA72E} 67.40} &
  {\ul 0.6680} &
  0.0034 &
  {\color[HTML]{4EA72E} 20.23} &
  \textbf{0.8705} &
  0.0034 &
  \cellcolor[HTML]{F2F2F2}{\color[HTML]{4EA72E} 5.72} &
  {\ul 0.7527} &
  {\color[HTML]{4EA72E} 24.84} \\
\rowcolor[HTML]{EFEFEF} 
+LoRA GRPO (semantic reward) &
  {\ul 0.7017} &
  0.0016 &
  {\color[HTML]{4EA72E} 63.23} &
  \textbf{0.7214} &
  0.0041 &
  {\color[HTML]{4EA72E} 29.84} &
  {\ul 0.8554} &
  0.0021 &
  \cellcolor[HTML]{F2F2F2}{\color[HTML]{4EA72E} 3.89} &
  \textbf{0.7595} &
  {\color[HTML]{4EA72E} 25.96} \\ \midrule \midrule
\rowcolor[HTML]{FFFFFF} 
+LoRA GRPO (semantic + cov/con rewards) &
  0.6912 &
  0.0024 &
  {\color[HTML]{4EA72E} 60.77} &
  0.6109 &
  0.0049 &
  {\color[HTML]{4EA72E} 9.95} &
  0.8529 &
  0.0032 &
  {\color[HTML]{4EA72E} 3.58} &
  0.7183 &
  {\color[HTML]{4EA72E} 19.13} \\
\rowcolor[HTML]{FFFFFF} 
+LoRA GRPO (w/o answer reward) &
  0.6704 &
  0.0022 &
  {\color[HTML]{4EA72E} 55.95} &
  0.5484 &
  0.0042 &
  {\color[HTML]{FF0000} -1.29} &
  0.8037 &
  0.0086 &
  {\color[HTML]{FF0000} -2.39} &
  0.6742 &
  {\color[HTML]{4EA72E} 11.82} \\ \midrule
\rowcolor[HTML]{FFFFFF} 
\multicolumn{12}{c}{\cellcolor[HTML]{FFFFFF}\textbf{Nitibench-Tax (Out-of-Domain)}} \\ \midrule
\rowcolor[HTML]{D9D9D9} 
openthaigpt1.5-qwen2.5-7b-instruct &
  0.1850 &
  0.0247 &
   &
  {\ul 0.3367} &
  0.0519 &
   &
  0.5400 &
  0.0849 &
   &
  0.3539 &
   \\
\rowcolor[HTML]{EFEFEF} 
+LoRA SFT &
  0.1039 &
  0.0387 &
  {\color[HTML]{FF0000} -43.84} &
  0.3267 &
  0.0450 &
  {\color[HTML]{FF0000} -2.97} &
  {\ul 0.5800} &
  0.0283 &
  \cellcolor[HTML]{F2F2F2}{\color[HTML]{4EA72E} 7.41} &
  0.3368 &
  {\color[HTML]{FF0000} -4.81} \\
\rowcolor[HTML]{EFEFEF} 
+LoRA GRPO (cov/con reward) &
  {\ul 0.2085} &
  0.0328 &
  {\color[HTML]{4EA72E} 12.73} &
  \textbf{0.3667} &
  0.0205 &
  {\color[HTML]{4EA72E} 12.24} &
  0.5600 &
  0.0748 &
  \cellcolor[HTML]{F2F2F2}{\color[HTML]{4EA72E} 3.70} &
  \textbf{0.3784} &
  {\color[HTML]{4EA72E} 6.93} \\
\rowcolor[HTML]{EFEFEF} 
+LoRA GRPO (semantic reward) &
  \textbf{0.2482} &
  0.0054 &
  {\color[HTML]{4EA72E} 34.16} &
  0.2500 &
  0.0424 &
  {\color[HTML]{FF0000} -25.74} &
  \textbf{0.6000} &
  0.0490 &
  \cellcolor[HTML]{F2F2F2}{\color[HTML]{4EA72E} 11.11} &
  {\ul 0.3661} &
  {\color[HTML]{4EA72E} 3.44} \\ \midrule \midrule
\rowcolor[HTML]{FFFFFF} 
+LoRA GRPO (semantic + cov/con rewards) &
  0.1830 &
  0.0048 &
  {\color[HTML]{FF0000} -1.04} &
  0.3067 &
  0.3682 &
  {\color[HTML]{FF0000} -8.91} &
  0.5267 &
  0.0499 &
  {\color[HTML]{FF0000} -2.47} &
  0.3388 &
  {\color[HTML]{FF0000} -4.26} \\
\rowcolor[HTML]{FFFFFF} 
+LoRA GRPO (w/o answer reward) &
  0.1662 &
  0.0090 &
  {\color[HTML]{FF0000} -10.16} &
  0.3133 &
  0.0125 &
  {\color[HTML]{FF0000} -6.93} &
  0.5333 &
  0.0189 &
  {\color[HTML]{FF0000} -1.23} &
  0.3376 &
  {\color[HTML]{FF0000} -4.60} \\ \bottomrule
\end{tabular}%
}
\caption{Additional results for OpenThaiGPT1.5-7B-Instruct on Nitibench-CCL and Nitibench-Tax. Compares LoRA GRPO performance using combined semantic and coverage/consistency (semantic + cov/con) rewards vs LoRA GRPO without any answer-specific reward (w/o answer reward).}
\label{tab:ablation_table}
\end{table*}

%% file: 6-ablation-study.tex
\section{Discussion}
\label{sec:ablation_study}

To understand reward contributions, we performed ablations on OpenThaiGPT1.5-7B (see Table \ref{tab:ablation_table}), comparing our main GRPO variants against configurations using: (1) combined semantic and coverage/consistency rewards ('semantic + cov/con'), and (2) only citation-related rewards ('w/o answer reward').

\subsection{Impact of Combining Answer Rewards}

\paragraph{Combining both rewards resulted in a worse performance gain.} 
Interestingly, simply combining both the semantic similarity reward and the coverage/consistency rewards did not lead to the best performance. 
As shown in Table \ref{tab:ablation_table}, this configuration (\emph{semantic + cov/con rewards}) underperformed compared to using either the semantic reward or the coverage/consistency reward individually across most metrics, particularly in terms of Coverage and Consistency gains on the Nitibench-CCL set, and showed poor generalization on the Nitibench-Tax set.
This suggests naive summation creates balancing issues or negative interference, potentially requiring better reward scaling or normalization.

\subsection{Impact of Removing Answer Rewards}

\paragraph{Using only citation-only reward enhances LLM citation ability for in-domain, but doesn't improve the quality of the generated response.}
While in-domain Citation F1 improved over baseline (+56\%), Coverage and Consistency degraded below baseline levels.
This variant also performed worst among GRPO configurations on CCL citation and failed to generalize on Tax (-10\% gain).
This strongly indicates that \textbf{generation quality aspects are coupled}; optimizing citations alone harms overall quality and generalization, demonstrating the need for answer quality rewards even to maximize citation performance within GRPO.

%% file: 7-discussion.tex
\begin{figure}[]
    \centering
    \includegraphics[scale=0.37]{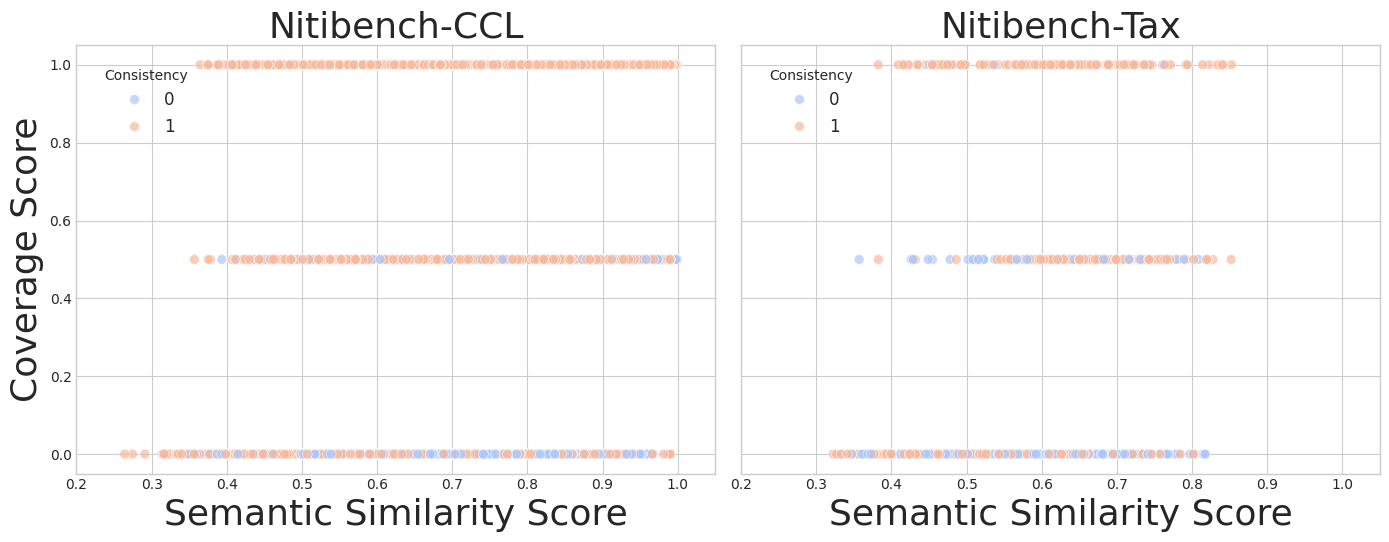}
    \caption{Semantic Similarity vs. Coverage scores, colored by Consistency, on (a) Nitibench-CCL and (b) Nitibench-Tax. A positive trend between similarity and coverage is more evident on CCL than on Tax.}
\label{fig:scatter_sim_cov_con}
\end{figure}

\begin{figure}[]
    \centering
    \includegraphics[scale=0.4]{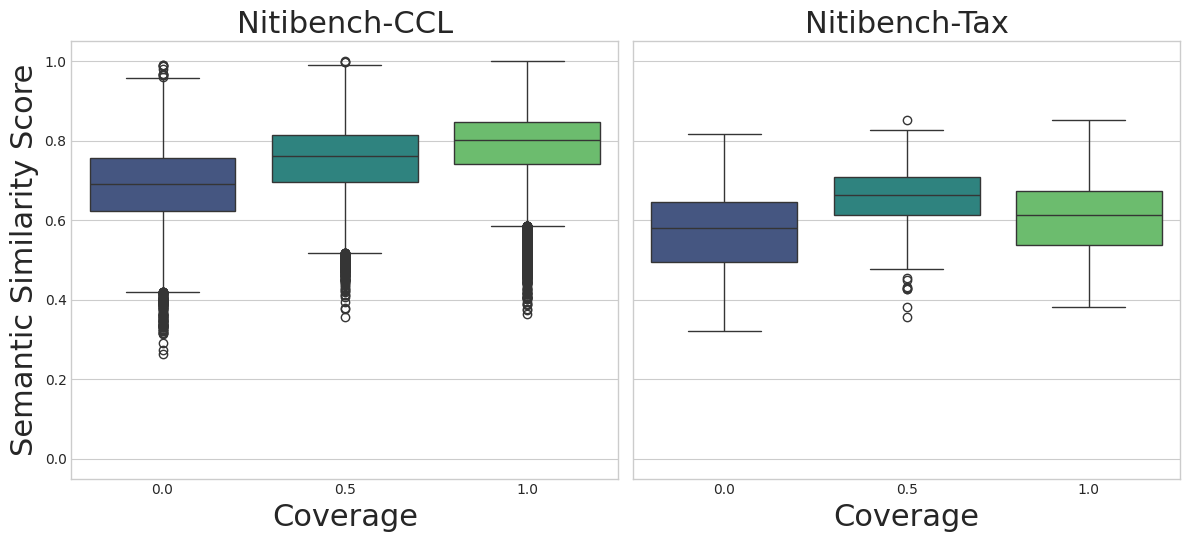}
 \caption{Semantic Similarity distributions by Coverage score level on (a) Nitibench-CCL and (b) Nitibench-Tax. Median similarity tends to increase with coverage on CCL, a trend not observed on Tax.}
\label{fig:boxplot_sim_coverage}

\end{figure}

\begin{figure}[]
    \centering
    \includegraphics[scale=0.4]{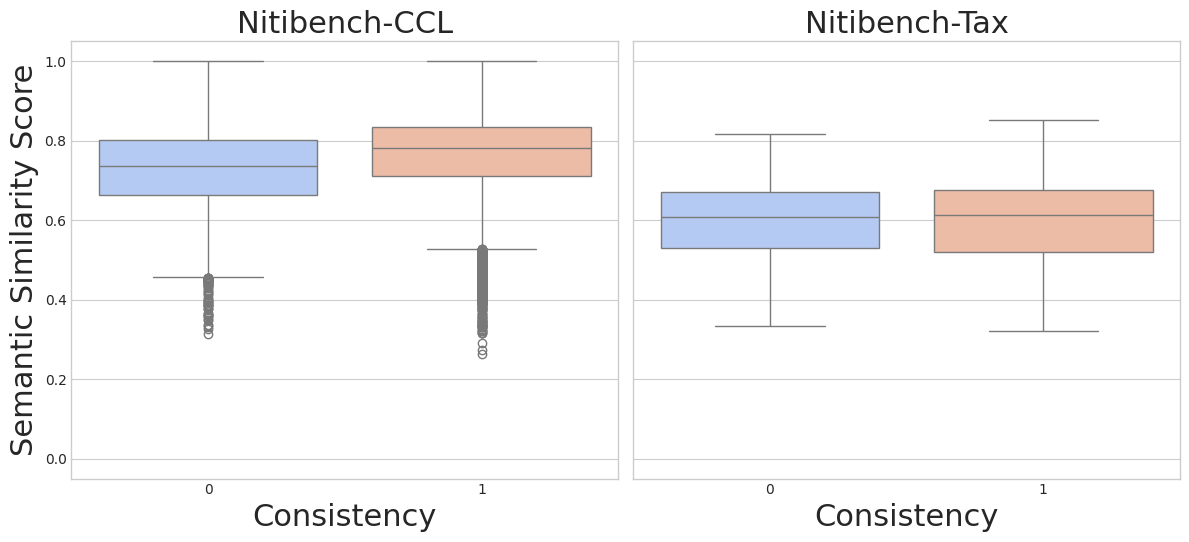}
\caption{Semantic Similarity distributions by Consistency score on (a) Nitibench-CCL and (b) Nitibench-Tax. Consistent answers on CCL show higher similarity; this distinction is less clear on Tax.}
\label{fig:boxplot_sim_consistency}
\end{figure}


\subsection{Correlation of Semantic Similarity with Coverage and Consistency}
\label{sec:correlation_analysis}

We investigated using BGE-M3 semantic similarity as an efficient proxy reward for answer quality during GRPO, avoiding costly judges like Qwen2.5-72B-Instruct for online training. To validate this proxy, we analyzed its correlation with ground-truth Coverage and Consistency scores (determined by an offline judge) on both Nitibench test sets in Figures \ref{fig:scatter_sim_cov_con}, \ref{fig:boxplot_sim_coverage}, and \ref{fig:boxplot_sim_consistency}.

For \textbf{Nitibench-CCL}, we observed a noticeable positive correlation. Higher semantic similarity generally aligns with higher Coverage and Consistency scores, as seen in both scatter and box plots. This suggests semantic similarity provides a meaningful, though imperfect, signal for answer quality on this simpler, in-domain dataset, supporting its use as a proxy reward here.

Conversely, for the more complex \textbf{Nitibench-Tax}, semantic similarity showed negligible correlation with Coverage or Consistency. The scatter plot lacked clear trends, and box plots revealed largely overlapping distributions for semantic similarity across different quality levels.

This contrast demonstrates that the utility of semantic similarity as a reward proxy is highly context-dependent. While adequate for simpler tasks (Nitibench-CCL), it fails to capture crucial aspects of correctness and factual consistency on complex reasoning tasks requiring synthesis (Nitibench-Tax), where semantic overlap alone is insufficient. The limitations of this efficient proxy become apparent on harder generalization problems. Appendix \ref{appendix:tax_complexity} provides a detailed comparison highlighting the increased complexity of Nitibench-Tax relative to Nitibench-CCL.

\subsection{Impact of Citation and Answer Position in Output Format}
\label{appendix:citation_position}

The standard output format used in our experiments follows the structure: reasoning -> answer -> citation (as in Appendix \ref{appendix:output_format}), where the model first provides its reasoning, then the synthesized answer, and finally the supporting citations. To investigate whether the position of the citation block relative to the answer block influences performance, we conducted an additional experiment.

We modified the target output structure to: reasoning -> citation -> answer, placing the citation block immediately after the reasoning and before the final answer. We then retrained the OpenThaiGPT1.5-7B-Instruct model using the GRPO (semantic reward) configuration with this modified "citation-first" target format. All other training parameters remained identical to the corresponding main experiment run.

The results of this comparison are presented in Table \ref{tab:citation_position_table}. The data clearly indicates that altering the standard format to place citations before the answer consistently resulted in \textbf{lower performance across nearly all metrics} on both the Nitibench-CCL and Nitibench-Tax datasets compared to the default format, where citations appear last. Notably, Citation F1, Coverage, and the overall Joint Score decreased in the "citation-first" configuration. On the challenging Nitibench-Tax set, this format led to performance even worse than the baseline in terms of Joint Score (-3.24\% gain).

While the exact reasons require deeper analysis, this finding suggests that the default structure (reasoning -> answer -> citation) may provide a more natural or effective flow for the model during generation and training. It's possible that generating the answer text first helps the model consolidate the information needed before explicitly listing the supporting citations. Regardless, based on these results, maintaining the structure with the citation block at the end appears preferable for achieving optimal performance with our GRPO approach.

\input{table/citation_position_table}

\subsection{Efficiency of Reward Signal Proxies}
\label{appendix:efficiency}
\begin{figure}[!h]
    \centering
    \includegraphics[scale=0.3]{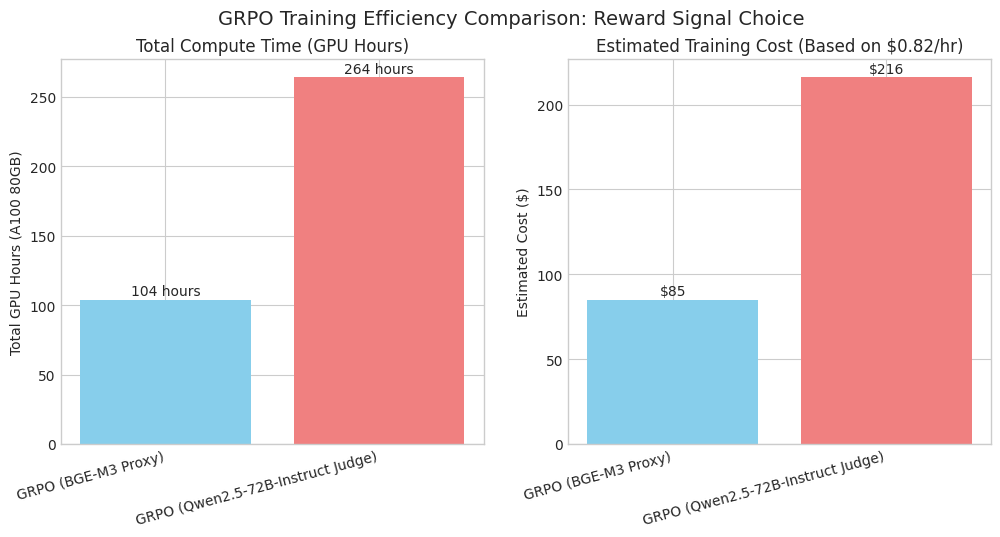}
    \caption{Comparison of total GPU hours and estimated training cost for GRPO variants using different answer reward signals.}
    \label{fig:efficiency_comparison}
\end{figure}

The practicality of RL hinges on reward computation efficiency. 
We observed a stark difference between using BGE-M3 semantic similarity versus the large Qwen2.5-72B-Instruct judge for coverage/consistency rewards. 
As shown in Figure \ref{fig:efficiency_comparison}, the BGE-M3 approach required significantly fewer resources per GRPO policy training: \textbf{104 total GPU-hours} (1x A100 80GB for 4 days 8 hours), costing approximately \textbf{\$85}. 
In contrast, using the Qwen2.5-72B-Instruct judge demanded \textbf{264 GPU-hours} (2x A100 80GB for 5 days 12 hours) - nearly 2.5x the compute time - costing roughly \textbf{\$216}\footnote{Based on A100 80GB PCIE median rental cost of \$0.82/hr via \url{https://vast.ai/pricing/gpu/A100-PCIE} accessed April 2025.}. 
This setup was necessary because one GPU was dedicated solely to hosting the 72B judge model as an online reward server, while the other GPU handled the training process.

This large disparity arises because the BGE-M3 embedding calculation is fast, adding minimal latency to the RL loop, while Qwen2.5-72B-Instruct inference for each sample creates a major bottleneck, requiring more hardware and time. 
While a large judge might offer reward signals closer to final evaluation metrics, its computational cost significantly hinders online RL training. 
BGE-M3 semantic similarity, despite being a proxy, proves vastly more efficient. 
Its strong performance, especially in-domain, confirms its value as a cost-effective method for injecting an answer quality signal during GRPO training.

%% file: table/citation_position_table.tex
\begin{table*}[]
\centering
\resizebox{\textwidth}{!}{%
\begin{tabular}{@{}lccccccccccc@{}}
\toprule
\rowcolor[HTML]{FFFFFF} 
model &
  Citation F1 $\uparrow$ &
  SD &
  gains (\%) &
  Coverage $\uparrow$ &
  SD &
  gains (\%) &
  Consistency $\uparrow$ &
  SD &
  gains (\%) &
  Joint score &
  gains (\%) \\ \midrule
\rowcolor[HTML]{FFFFFF} 
\multicolumn{12}{c}{\cellcolor[HTML]{FFFFFF}\textbf{Nitibench-CCL}} \\ \midrule
\rowcolor[HTML]{D9D9D9} 
openthaigpt1.5-qwen2.5-7b-instruct &
  0.4299 &
  0.0048 &
   &
  0.5556 &
  0.0010 &
   &
  0.8234 &
  0.0048 &
   &
  0.6030 &
   \\
\rowcolor[HTML]{EFEFEF} 
+LoRA GRPO (semantic reward) &
  \textbf{0.7017} &
  0.0016 &
  {\color[HTML]{4EA72E} 63.23} &
  \textbf{0.7214} &
  0.0041 &
  {\color[HTML]{4EA72E} 29.84} &
  \textbf{0.8554} &
  0.0021 &
  \cellcolor[HTML]{F2F2F2}{\color[HTML]{4EA72E} 3.89} &
  \textbf{0.7595} &
  {\color[HTML]{4EA72E} 25.96} \\
\rowcolor[HTML]{FFFFFF} 
+LoRA GRPO (semantic reward, citation first) &
  {\ul 0.6545} &
  0.0044 &
  {\color[HTML]{4EA72E} 52.25} &
  {\ul 0.7065} &
  0.0053 &
  {\color[HTML]{4EA72E} 27.16} &
  {\ul 0.8528} &
  0.0028 &
  {\color[HTML]{4EA72E} 3.57} &
  0.7379 &
  {\color[HTML]{4EA72E} 22.39} \\ \midrule
\rowcolor[HTML]{FFFFFF} 
\multicolumn{12}{c}{\cellcolor[HTML]{FFFFFF}\textbf{Nitibench-Tax}} \\ \midrule
\rowcolor[HTML]{D9D9D9} 
openthaigpt1.5-qwen2.5-7b-instruct &
  0.1850 &
  0.0247 &
   &
  \textbf{0.3367} &
  0.0519 &
   &
  {\ul 0.5400} &
  0.0849 &
   &
  0.3539 &
   \\
\rowcolor[HTML]{F2F2F2} 
\cellcolor[HTML]{EFEFEF}+LoRA GRPO (semantic reward) &
  \textbf{0.2482} &
  0.0054 &
  {\color[HTML]{4EA72E} 34.16} &
  0.2500 &
  0.0424 &
  {\color[HTML]{FF0000} -25.74} &
  \textbf{0.6000} &
  0.0490 &
  {\color[HTML]{4EA72E} 11.11} &
  {\ul 0.3661} &
  {\color[HTML]{4EA72E} 3.44} \\
\rowcolor[HTML]{FFFFFF} 
+LoRA GRPO (semantic reward, citation first) &
  {\ul 0.2172} &
  0.0146 &
  {\color[HTML]{4EA72E} 17.43} &
  {\ul 0.2768} &
  0.0026 &
  {\color[HTML]{FF0000} -17.79} &
  0.5333 &
  0.0411 &
  {\color[HTML]{FF0000} -1.24} &
  0.3424 &
  {\color[HTML]{FF0000} -3.24} \\ \bottomrule
\end{tabular}%
}
\caption{Comparison of GRPO (semantic reward) performance on OpenThaiGPT1.5-7B using the default output format (reasoning->answer->citation) versus a modified format placing citations before the answer (reasoning->citation->answer).}
\label{tab:citation_position_table}
\end{table*}

%% file: 8-conclusion.tex
\section{Conclusion}
\label{sec:conclusion}

This work demonstrates that Group Relative Policy Optimization (GRPO) substantially outperforms Instruction Tuning (IT) for Thai Legal QA, yielding significant citation accuracy gains on the in-domain Nitibench-CCL dataset. Critically, GRPO uniquely improved generalization performance on the complex Nitibench-Tax set when applied to Thai-aligned base models, contrasting with SFT's consistent degradation, thus showing potential for fostering more robust legal reasoning.

Our investigation into reward functions revealed practical trade-offs. An efficient semantic similarity proxy reward proved effective and cost-efficient for in-domain tasks (Nitibench-CCL), correlating reasonably well with quality metrics. However, this correlation failed on the harder Nitibench-Tax task, limiting its utility for complex generalization. Conversely, using a large judge model (Qwen2.5-72B-Instruct) provided a potentially more accurate signal but incurred significantly higher computational costs (>2.5x). Ablation studies confirmed the necessity of combining answer quality rewards with citation rewards for best results.

In conclusion, GRPO is a promising and efficient fine-tuning approach for specialized domains. Its success depends on synergizing the RL method, base model capabilities, and reward design. Future work should focus on developing reward mechanisms that balance accuracy for complex reasoning with practical training efficiency, improving upon current proxies without resorting to prohibitively expensive judge models.

%% file: 9-limitation.tex
\section*{Limitations}
\label{sec:limitations}
While this study provides valuable insights into applying GRPO for Thai Legal QA, we acknowledge certain limitations primarily stemming from constraints on computational resources and time during the experimental phase.

First, our exploration of combining different reward signals for answer quality. 
Specifically, the semantic similarity reward and the coverage/consistency rewards from the Qwen2.5-72B-Instruct judge were limited. 
The ablation study used a naive summation without tuning, which underperformed relative to individual signals. 
Due to resource constraints, we were unable to explore alternative reward calibration strategies such as weighting, normalization, or learning rate adjustments. 
A well-tuned combination may offer synergistic benefits, but this remains unexplored.

Second, our experiments focused exclusively on applying GRPO to models that had already undergone instruction tuning. 
%
We did not evaluate applying GRPO directly to base models (e.g., \cite{deepseekr1}). 
Investigating its effect from different model initialization states may yield further insights, but it was beyond our current scope.

Third, we used the standard GRPO algorithm as described by \cite{grpo}. 
While conducting our experiments, an improved variant named "Dr. GRPO" (Done Right GRPO) was proposed \cite{dr_grpo}, specifically designed to address optimization biases present in the original GRPO formulation, particularly those related to response length normalization, which can affect token efficiency. 
Due to the timing of its release relative to our experimental runs and resource limitations, we were unable to incorporate Dr. GRPO into our comparisons. 
We acknowledge the potential biases in standard GRPO identified by \cite{dr_grpo} and recognize that employing Dr. GRPO might yield different results, particularly regarding token efficiency and potentially performance dynamics.

These limitations reflect the demonstrative nature of this study, which aims to assess the potential of GRPO for citation-sensitive legal QA under domain-specific constraints. 
Addressing them may deepen our understanding of GRPO’s behavior in legal settings and inform strategies for best utilizing it in practice.

%% file: 10-appendix.tex
\appendix

\section{Hyperparameters}
\label{appendix:hyperparameters}

\subsection{Training Hyperparameters}
\label{appendix:training_params}

This section details the key hyperparameters used for Instruction Tuning Fine-tuning (IT) and Group Relative Policy Optimization (GRPO) training procedures described in Section \ref{sec:experimental_setup}. Common parameters related to LoRA configuration, precision, optimizer betas, and data handling were kept consistent where applicable.

\begin{table}[!h]
\resizebox{0.5\textwidth}{!}{%
    \centering
    \begin{tabular}{l|l|l}
    \hline
    \textbf{Hyperparameter} & \textbf{GRPO Value} & \textbf{IT Value} \\
    \hline
    Learning Rate (lr) & 5.00E-06 & 1.00E-05 \\
    LR Scheduler Type & \texttt{constant\_with\_warmup} & \texttt{cosine} \\
    Max Gradient Norm & 0.2 & 1.0 \\
    Epochs & 1 & 3 \\
    Rollout Batch Size & 10 & N/A \\
    SFT Batch Size & N/A & 4 \\
    Max Prompt Length & 8192 & 8192 \\
    Max Completion Length & 2048 & 2048 \\
    LoRA Rank (\(r\)) & 256 & 256 \\
    Precision & bfloat16 & bfloat16 \\
    Retrieval Top-k & 10 & 10 \\ 
    Gradient Accumulation Steps & 1 & 1 \\
    Weight Decay & 0.1 & 0.1 \\
    Warmup Ratio & 0.1 & 0.1 \\
    Adam Beta1 & 0.9 & 0.9 \\
    Adam Beta2 & 0.99 & 0.99 \\
    \hline
    \end{tabular}
}
\caption{Comparison of Key Hyperparameters for SFT and GRPO Training.}
\label{tab:hyperparameters}
\end{table}

\section{Input and Output Formats}
\label{appendix:input_output_format}

This section provides concrete examples of the input prompt structure fed to the models and the target output format used during fine-tuning (both SFT and GRPO), complementing the description in Section \ref{sec:experimental_setup}.

\subsection{Example Input Prompt Structure}
The following illustrates the format of the input provided to the models. This example assumes the context retrieval resulted in \(k=5\) relevant sections after length management. The \texttt{\{context\}} placeholder represents the actual text content of the corresponding legal section. The \texttt{<law\_code>} tags contain unique integer identifiers assigned to each distinct legal section within our corpus; these identifiers are used as keys and do not necessarily correspond to official statutory section numbers.

\begin{lstlisting}[numbers=left,xleftmargin=5mm]
What is the difference between financial institution business and financial business?

Relevant sections
<law_code>1</law_code><context>...</context>
<law_code>2</law_code><context>...</context>
<law_code>3</law_code><context>...</context>
<law_code>4</law_code><context>...</context>
<law_code>5</law_code><context>...</context>
\end{lstlisting}

\subsection{Example Target Output Structure}
\label{appendix:output_format}
The models were trained to generate outputs adhering to the following XML-like structure. This format separates the reasoning process, the final answer, and the cited sources.

\begin{lstlisting}[numbers=left,xleftmargin=5mm]
<reasoning>
The laws related to the method for director resignation are ...
</reasoning>
<answer>
According to Section 1153/1 of the Civil and Commercial Code and ...
</answer>
<citation>
<law_code>2</law_code>
<law_code>5</law_code>
</citation>
\end{lstlisting}

\textbf{Note:} The \texttt{<reasoning>} block contains the model's generated explanation or thought process. The \texttt{<answer>} block contains the final synthesized answer to the query. The \texttt{<citation>} block lists the \texttt{<law\_code>} identifiers that the model cites as sources for its answer. During IT, this structure represents the target output. During GRPO, adherence to this format and the correctness of the content within the tags (\texttt{<answer>} and \texttt{<citation>}) are evaluated by the reward functions.

\subsection{Inferencing Hyperparameters}
\label{appendix:inferencing_params}

This section details the hyperparameters used during the inference phase to generate the model outputs for the final evaluation presented in Section \ref{sec:results}. These settings were applied consistently across all model configurations (Baseline, SFT, GRPO) when evaluating on the Nitibench-CCL and Nitibench-Tax test sets using vLLM \cite{vllm}. The following parameters were used for text generation:

\begin{description}
    \item[Generation Seeds:] Inference was repeated three times for each model configuration using the following distinct random seeds: \texttt{69420}, \texttt{69421}, and \texttt{69422}. The final reported metrics are the mean and standard deviation across these three runs (as detailed in Section \ref{sec:inference_aggregation}).
    \item[Retrieval Top-k:] Set to \texttt{10}, same as the \texttt{Retrieval Top-k} in the training hyperparameter.
    \item[Temperature:] Set to \texttt{1.0} for standard diversity in the output.
\end{description}

\section{Evaluation of Qwen-72B as an Automated Judge}
\label{appendix:human_agreement_qwen72b}

To assess the viability of using Qwen2.5-72B-Instruct as an online judge for generating Coverage and Consistency rewards in GRPO (Section \ref{sec:grpo_rewards}), we compared its judgment reliability against \texttt{gpt-4o-2024-08-06} on the Nitibench-CCL dataset, as it achieved the highest performance among judges evaluated in the original Nitibench paper \cite{nitibench}. We follow Nitibench's decoding hyperparameters: \texttt{temperature = 0.5}, \texttt{seed = 69420}, and \texttt{max\_completion\_tokens = 2048}.

As shown in Table \ref{tab:judge_performance}, Qwen-72B achieved high reliability, closely matching GPT-4o. For \textbf{Coverage}, Qwen-72B reached an F1-score of 0.84 (vs. 0.88 for GPT-4o), and for \textbf{Consistency}, it scored 0.97 (vs. 0.98 for GPT-4o). These results demonstrate that Qwen2.5-72B-Instruct functions as a reliable automated judge for these metrics on this dataset, validating its use for providing sufficiently accurate reward signals during GRPO training as an alternative to external API calls.

\begin{table}[!h]
\centering
\resizebox{0.5\textwidth}{!}{%
\begin{tabular}{@{}llllll@{}}
\toprule
\textbf{Model} & \textbf{Metric} & \textbf{Precision} & \textbf{Recall} & \textbf{F1-score} & \textbf{Support} \\ \midrule
\multicolumn{6}{c}{Nitibench-CCL}                                        \\ \midrule
{gpt-4o-2024-08-06} & Coverage    & .88 & .88 & .88 & 200 \\
                                   & Consistency & .98 & .97 & .98 & 150 \\ \midrule
Qwen2.5-72B-Instruct               & Coverage    & .85 & .83 & .84 & 200 \\
                                   & Consistency & .98 & .97 & .97 & 150 \\ \bottomrule
\end{tabular}%
}
\caption{Performance comparison of GPT-4o (\texttt{gpt-4o-2024-08-06}) and Qwen2.5-72B-Instruct as automated judges for Coverage and Consistency metrics on the Nitibench-CCL dataset.}
\label{tab:judge_performance}
\end{table}

\section{Complexity of Nitibench-Tax over Nitibench-CCL}
\label{appendix:tax_complexity}

\begin{figure}[!h]
    \centering
    \includegraphics[scale=0.25]{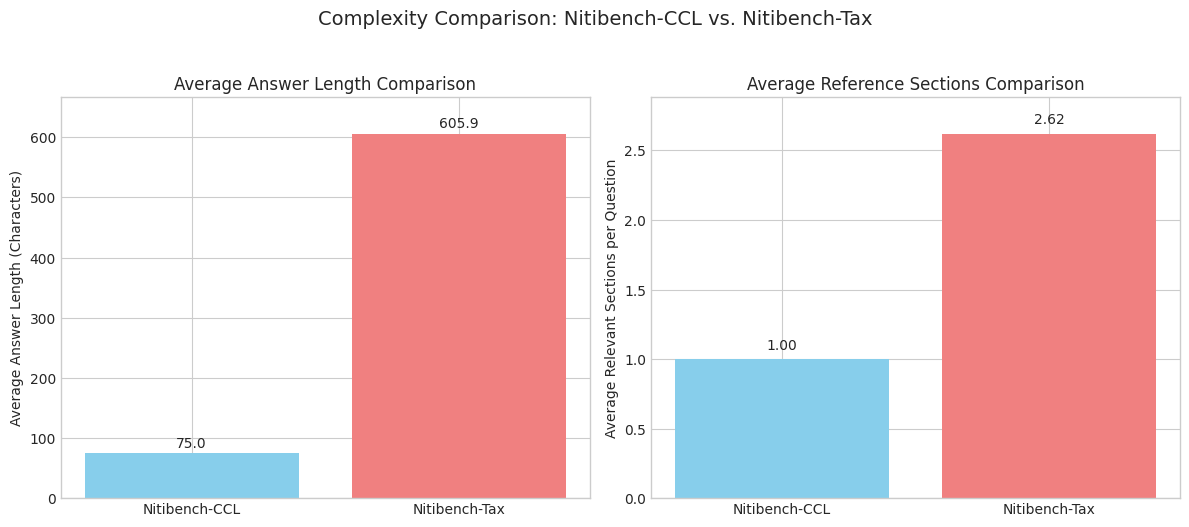}
    \caption{Complexity Comparison of Nitibench-CCL vs. Nitibench-Tax.}
    \label{fig:tax_ccl_compare}
\end{figure}

While both Nitibench-CCL and Nitibench-Tax evaluate Thai Legal QA, the Nitibench-Tax dataset presents a significantly more complex challenge, designed specifically to test model generalization and deeper reasoning capabilities (see Figure \ref{fig:tax_ccl_compare} for answer length and section per answer comparison). This difference stems from several key aspects of their origin and structure:

\begin{enumerate}
    \item \textbf{Dataset Origin and Curation:}
        \begin{itemize}
            \item \textbf{Nitibench-CCL:} This dataset was curated manually by legal experts who crafted question-answer pairs primarily based on single, specific legal sections from a defined corpus of 35 financial laws. The process involved a two-tiered expert review to ensure quality. While its corresponding training data (from WangchanX-Legal-ThaiCCL-RAG\footnote{https://huggingface.co/datasets/airesearch/WangchanX-Legal-ThaiCCL-RAG}) could be multi-label due to semi-automated generation, the test set used for evaluation predominantly consists of single-label instances.
            \item \textbf{Nitibench-Tax:} This dataset originates from real-world tax rulings scraped directly from the Thai Revenue Department's official website\footnote{https://www.rd.go.th} (cases from 2021 onwards). These represent authentic inquiries and official responses, reflecting the complexity of actual tax law application. The curation involved extracting relevant cited sections and condensing the official responses using an LLM, after filtering out non-interpretive cases.
        \end{itemize}
        The use of real, official rulings in Nitibench-Tax inherently introduces more complex scenarios and language compared to the expert-crafted, typically single-provision-focused questions in the Nitibench-CCL test set.

    \item \textbf{Answer Length and Complexity:}
        The complexity difference is reflected in the average length of the ground-truth answers (after condensation). The average answer length in \textbf{Nitibench-CCL is approximately 75 characters}, whereas in \textbf{Nitibench-Tax, it is roughly 606 characters} - over eight times longer on average. This suggests that Tax answers inherently require significantly more detail and potentially cover more sub-points derived from the underlying complex rulings.

    \item \textbf{Multi-Label Nature (Sections per Answer):}
        This is a critical quantitative differentiator. The Nitibench-CCL test set is explicitly single-label, with an average of \textbf{1 ground-truth relevant legal section} per question. In contrast, Nitibench-Tax is inherently multi-label, with an average of \textbf{2.62 relevant sections} per case. This requires models not just to identify relevant sections but to synthesize information and reason across multiple legal provisions simultaneously, significantly increasing the reasoning complexity compared to the single-label focus of CCL.
\end{enumerate}

In summary, the combination of using real-world, complex tax rulings as source material and its inherent multi-label requirement (demanding reasoning across multiple sections) makes Nitibench-Tax a substantially harder benchmark than Nitibench-CCL for evaluating advanced legal reasoning and generalization abilities.